\documentclass[letterpaper, 10 pt, conference]{ieeeconf}  % Comment this line out if you need a4paper

\IEEEoverridecommandlockouts                              % This command is only needed if 
\usepackage[T1]{fontenc}
\usepackage{preamble}

\title{\LARGE \bf
Realm: Real-Time Line-of-Sight Maintenance in Multi-Robot Navigation with Unknown Obstacles
}

\author{Ruofei Bai$^{1, 2}$, Shenghai Yuan$^{1}$, Kun Li$^{3}$, Hongliang Guo$^{4}$, Wei-Yun Yau$^{2}$, Lihua Xie$^{1}$, \emph{Fellow}, \emph{IEEE}% <-this % stops a space
\thanks{$^{1}$Ruofei Bai, Shenghai Yuan and Lihua Xie are with the School of
Electrical and Electronic Engineering, Nanyang Technological University, Singapore 639798
        {\tt\small \{ruofei001, shyuan, elhxie\}@ntu.edu.sg}}%
\thanks{$^{2}$Ruofei Bai and Wei-Yun Yau are with the Institute for Infocomm Research (I2R), Agency for Science, Technology and Research (A*STAR), Singapore 138632
        {\tt\small \{stubair, wyyau\}@i2r.a-star.edu.sg}}%
\thanks{$^{3}$Kun Li is with the School of
Automation, Chongqing University, China 400044
        {\tt\small likun@cqu.edu.cn}}%
\thanks{$^{4}$Hongliang Guo is with College of
Computer Science, Sichuan University, China 610064
        {\tt\small guohongliang@scu.edu.cn}}%
}

\begin{document}

\maketitle
\thispagestyle{empty}
% \pagestyle{empty}
% 启用页码显示
\pagestyle{plain}

%%%%%%%%%%%%%%%%%%%%%%%%%%%%%%%%%%%%%%%%%%%%%%%%%%%%%%%%%%%%%%%%%%%%%%%%%%%%%%%%
\begin{abstract}
Multi-robot navigation in complex environments relies on inter-robot communication and mutual observations for coordination and situational awareness.
This paper studies the multi-robot navigation problem in unknown environments with line-of-sight (LoS) connectivity constraints.
While previous works are limited to known environment models to derive the LoS constraints, this paper eliminates such requirements by directly formulating the LoS constraints between robots from their real-time point cloud measurements, leveraging point cloud visibility analysis techniques.
We propose a novel LoS-distance metric to quantify both the urgency and sensitivity of losing LoS between robots considering potential robot movements.
Moreover, to address the imbalanced urgency of losing LoS between two robots, we design a fusion function to capture the overall urgency while generating gradients that facilitate robots' collaborative movement to maintain LoS.
The LoS constraints are encoded into a potential function that preserves the positivity of the Fiedler eigenvalue of the robots' network graph to ensure connectivity.
Finally, we establish a LoS-constrained exploration framework that integrates the proposed connectivity controller.
We showcase its applications in multi-robot exploration in complex unknown environments, where robots can always maintain the LoS connectivity through distributed sensing and communication, while collaboratively mapping the unknown environment.
The implementations are open-sourced at \url{https://github.com/bairuofei/LoS_constrained_navigation}.

\end{abstract}

%%%%%%%%%%%%%%%%%%%%%%%%%%%%%%%%%%%%%%%%%%%%%%%%%%%%%%%%%%%%%%%%%%%%%%%%%%%%%%%%

\section{Introduction}

Multi-robot navigation holds various applications in search and rescue~\cite{tian2020search}, area inspection~\cite{10591842}, and autonomous exploration~\cite{10577228}, etc.
While robots usually rely on inter-robot communication for timely coordination, its reliability can be compromised in practice due to the power limitation of robots' onboard communication devices and the obstruction of obstacles in environments.
Moreover, mutual observations between robots can also be critical for real-time situational awareness when robots are deployed in dangerous environments. 
Most existing works assume range-based communication models~\cite{shi_CommunicationAwareMultirobot_2021, cao2023representation, tan2024ir, varadharajan2020swarm}, while omitting the presence of obstacles that can obstruct Line-of-Sight (LoS) between robots and degrade radio signals, leading to disrupted communication and observation within the multi-robot system~\cite{liu_RelativeLocalizationEstimation_2023}.
% However, obstacles in environments can weaken the communication signal and block the line-of-sight between robots, interrupt their mutual observations, causing disconnection among members in a multi-robot system.
To support reliable communication and coordination of multi-robot systems, one challenging problem is to maintain the LoS-connectivity between robots in the presence of obstacles while performing external navigation tasks~\cite{amigoni_MultirobotExploration_2017}.

Existing works address this problem by assuming known environment models, based on which the robot connectivity can be maintained in either a continuous or a discrete manner~\cite{amigoni_MultirobotExploration_2017, xia_RELINKRealTime_2023, robuffogiordano_PassivitybasedDecentralized_2013, nestmeyer_DecentralizedSimultaneous_2017}.
The environment models help to formulate the potential functions to keep LoS between robots~\cite{robuffogiordano_PassivitybasedDecentralized_2013, amigoni_MultirobotExploration_2017}, define safe zones for robots that ensure connectivity~\cite{chen_MultiUAVDeployment_2023}, or check the connectivity between robots' target positions in discrete decision-making~\cite{shi_CommunicationAwareMultirobot_2021}.

However, challenges arise when robots are deployed in unknown environments, where unpredictable obstacles may hinder the LoS connectivity between robots.
Moreover, it is non-trivial to construct a consistent and concise obstacle point set among robots to describe the surrounding environment, which is usually required by traditional connectivity control methods~\cite{robuffogiordano_PassivitybasedDecentralized_2013}.

\begin{figure}[!t]
\centering
\includegraphics[width=\linewidth]{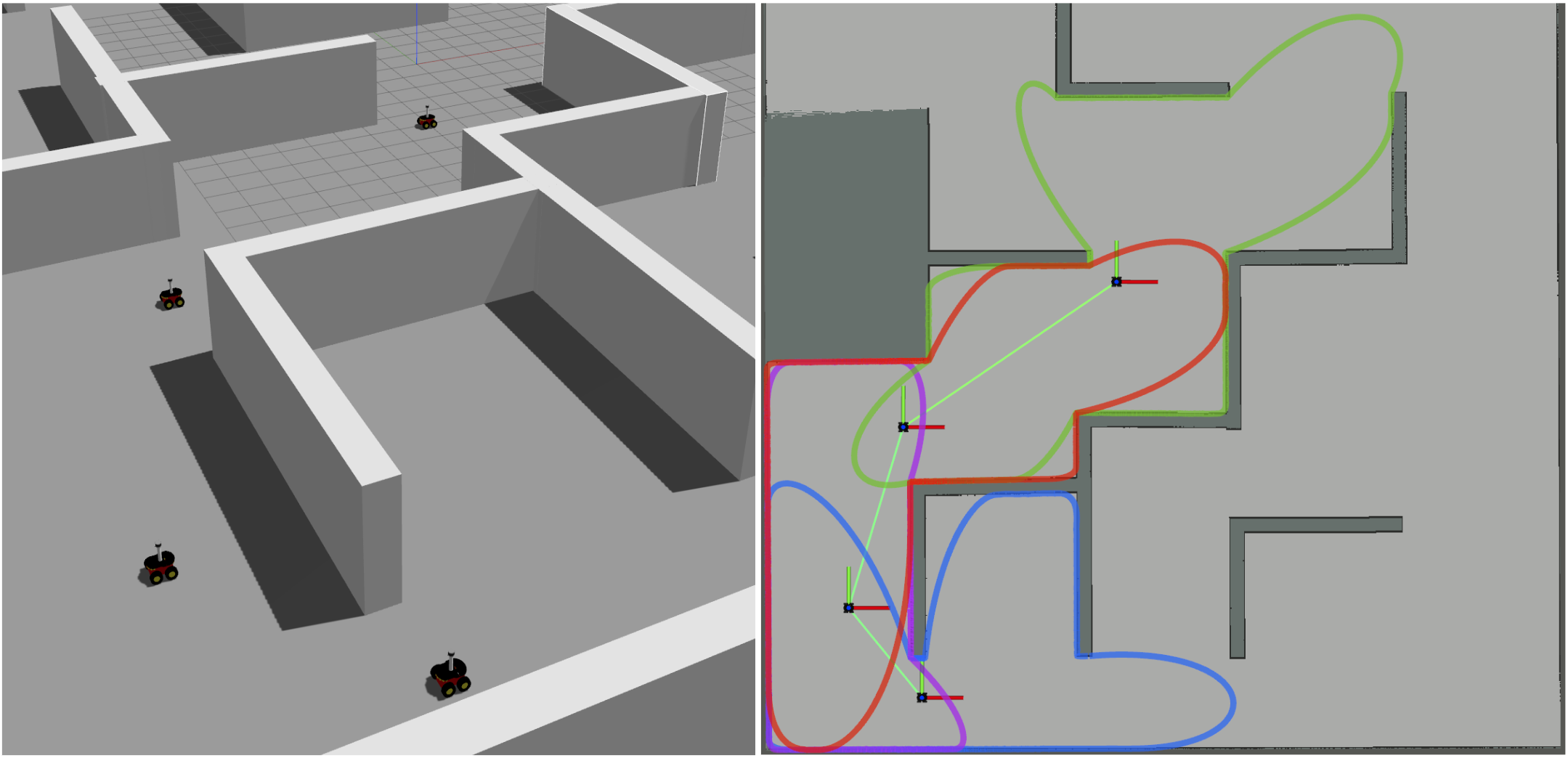}
\caption{Real-time LoS maintenance of four robots while exploring an unknown environment. Robots' visible regions (enclosed by curves with different colors) are constructed from their real-time point cloud measurements.
The connectivity between robots is shown by the light green edges. To maintain connectivity, each robot only needs local information from its one-hop neighbors.}
\label{fig_face}
\end{figure}

In this paper, we propose \textbf{Realm}, a \textbf{Rea}l-time \textbf{L}oS \textbf{m}aintenance method that can be applied in complex unknown environments without relying on prior environment models, as shown in Fig.~\ref{fig_face}.
% Different from existing works, our method formulates the LoS constraints between robots directly from their real-time point cloud measurements, eliminating the requirement of prior environmental models.
Our method can directly formulate the LoS constraints between robots based on their real-time point cloud measurements (from 2D/3D LiDARs, RGB-D cameras, etc), by adopting techniques in point cloud visibility analysis.
The LoS constraints are then encoded into a potential function to preserve the positivity of the Fiedler eigenvalue of the graph Laplacian matrix of robots' underlying topology, which ensures the connectivity between robots.
The control command of robots can be calculated distributively with only one-hop communication between its immediate neighbors, and allows time-varying connected topology of robots while performing external navigation tasks.
Our contributions are summarized as follows:

\begin{enumerate}
    \item We derive a novel LoS-distance metric that encodes both the urgency and sensitivity of losing LoS between robots, extending visibility analysis techniques from computer graphics to real-time formulation of inter-robot LoS constraints;
    \item We design an effective LoS-distance fusion function that fuses imbalanced LoS-distance into a weighted connectivity graph, facilitating robots' collaborative movement to maintain LoS;
    \item We establish a LoS-constrained multi-robot navigation framework that integrates multi-robot mapping, task assignment, and path planning.
\end{enumerate}
We demonstrate the applications of the proposed method in multi-robot exploration of complex unknown environments, where robots can successfully explore the entire environment while always ensuring LoS-connectivity.
Our implementations will be open-sourced at \url{https://github.com/bairuofei/LoS_constrained_navigation}.

\section{Related Works}

\subsection{Line-of-Sight-Constrained Multi-Robot Navigation}

Existing works considering connectivity maintenance can be categorized into continuous and discrete types~\cite{amigoni_MultirobotExploration_2017}.
Continuous connectivity requires robots to always maintain a connected network during their operations.
Giordano~\etal~propose a potential function-based method that guarantees connectivity by preserving the Fiedler eigenvalue of the underlying graph Laplacian to be positive~\cite{robuffogiordano_PassivitybasedDecentralized_2013, amigoni_MultirobotExploration_2017}.
In their work, the LoS constraints are captured by the distance from the closest obstacle point to the LoS segment joining two robots.
Alternatively, an analytical form of the LoS constraints is also derived in~\cite{yang_MinimallyConstrained_2023} by approximating the LoS segment with a minimum volume enclosing ellipsoid (MVEE).
% They further consider the estimation uncertainty in robot positions in~\cite{yang_DecentralizedMultirobot_2024}. 
Chen~\etal~also consider the LoS constraints in multi-UAV deployment~\cite{chen_MultiUAVDeployment_2023}, where the movements of two robots are restricted in a safe zone defined by separating hyperplanes of obstacles.

In contrast, discrete connectivity only requires robots to be connected at certain time steps~\cite{hollinger_MultirobotCoordination_2012}.
A typical procedure first determines a target connected topology for the robots based on spanning tree search, after which their positions are refined through local optimization to balance additional objectives such as travel distance or information gain~\cite{shi_CommunicationAwareMultirobot_2021, stump_VisibilitybasedDeployment_2011, dutta_MultirobotInformative_2019, xia_RELINKRealTime_2023}.
To ensure LoS constraints in the local optimization stage, Stump~\etal~apply polygonal decomposition of an environment, after which neighboring robots' positions are restricted in a common polyhedron to ensure LoS~\cite{stump_VisibilitybasedDeployment_2011}.
Xia~\etal~formulate the LoS constraints based on robots' visible regions described by star convex polytope~\cite{xia_RELINKRealTime_2023}.
However, as local optimization modifies robots' positions, the corresponding changes in their visible regions may lead to potential LoS loss after optimization.

While most existing works rely on known environment models described by either obstacle points~\cite{robuffogiordano_PassivitybasedDecentralized_2013, yang_MinimallyConstrained_2023} or occupancy maps~\cite{stump_VisibilitybasedDeployment_2011, xia_RELINKRealTime_2023},
few works have considered connectivity maintenance in unknown environments.
Li~\etal~propose a reinforcement learning-based approach that generates control commands based on the inputs of range sensor measurements and the positions of other robots, and does not rely on prior environmental information~\cite{li_DecentralizedGlobal_2022}.
However, the method has no guarantee of global connectivity.

\section{Preliminaries}

\subsection{Graph Laplacian and Generalized Graph Connectivity}

Given a graph $\G = \langle \V, \E \rangle $ with $N$ vertices, the graph Laplacian matrix of $\G$ is defined as $\Lp = D - A$, where $A\in \mathbb{R}^{N\times N}$ is the adjacency matrix with an element $A_{ij} = 0$ if $(i, j)\notin \E$, and $A_{ij} > 0$ otherwise; $D\in \mathbb{R}^{N\times N}$ is a diagonal degree matrix where $D_{ii} = \sum_{j = 1}^{N}A_{ij}$ and $D_{ij}= 0$ when $i\ne j$. 
The second-smallest eigenvalue $\lambda_2$ of $\Lp$, usually referred to as the Fiedler eigenvalue~\cite{fiedler1973algebraic} or connectivity eigenvalue, can be used to check the connectivity of the graph $\G$.
It holds the property that $\lambda_2 > 0$ if the graph $\G$ is connected, or $\lambda_2 = 0$ otherwise.

By encoding the satisfaction levels of different kinds of connectivity constraints between two robots into the weight of their corresponding edge, the Fiedler eigenvalue $\lambda_2$ of the weighted graph Laplacian can be used to reflect the \emph{generalized} graph connectivity~\cite{robuffogiordano_PassivitybasedDecentralized_2013}.
In this work, we also use this concept to ensure connectivity between robots by enforcing that the Fiedler eigenvalue $\lambda_2$ of the weighted graph Laplacian remains positive during navigation.

\subsection{Visible Region Construction from Point Cloud}
\label{sec_preliminary_visibility}

\begin{figure}[t]
\vspace{6pt}
\centering
\subfloat[]{\includegraphics[width=.44\linewidth]{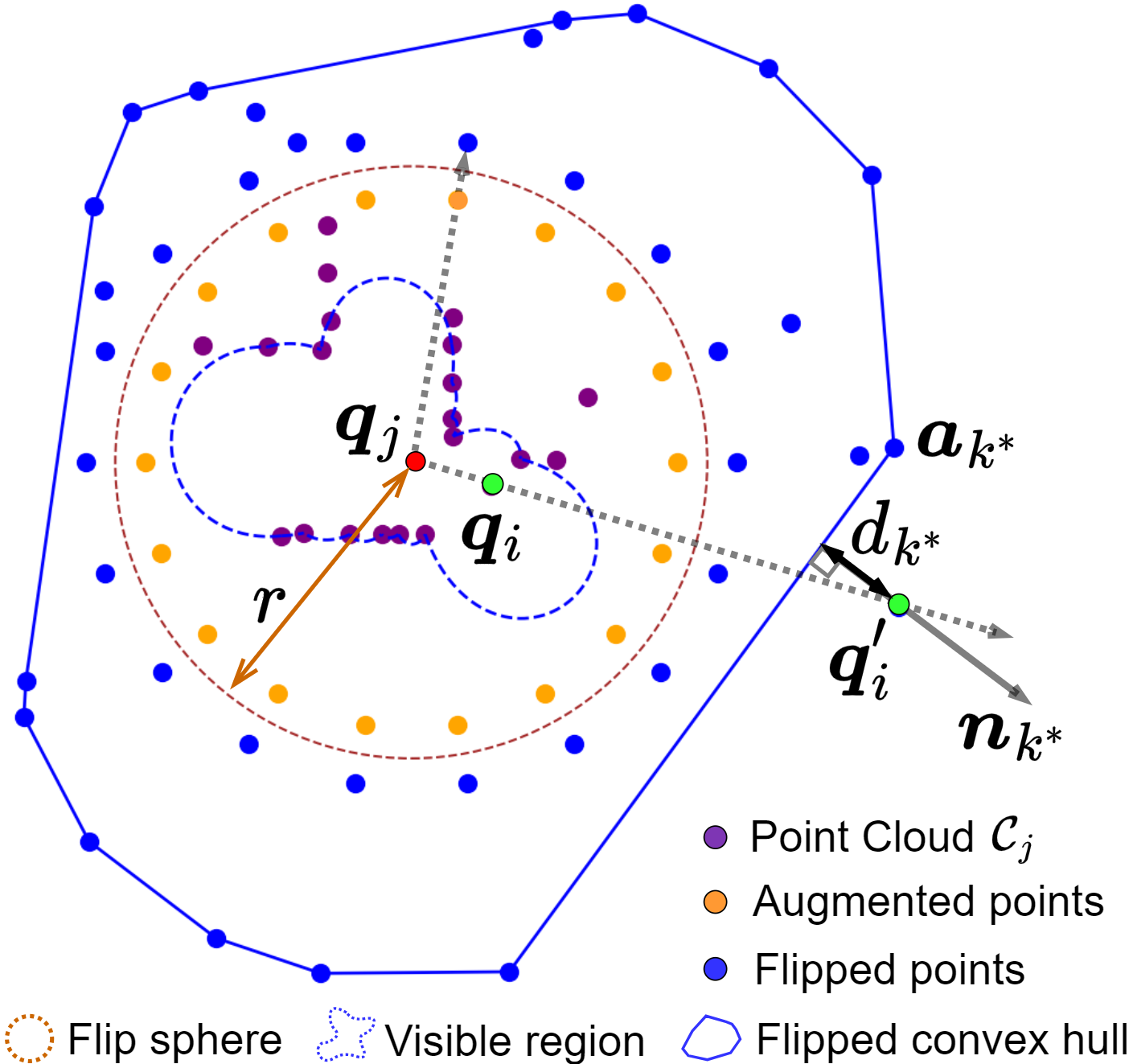}}\hspace{1pt}
\subfloat[]{\includegraphics[width=.54\linewidth]{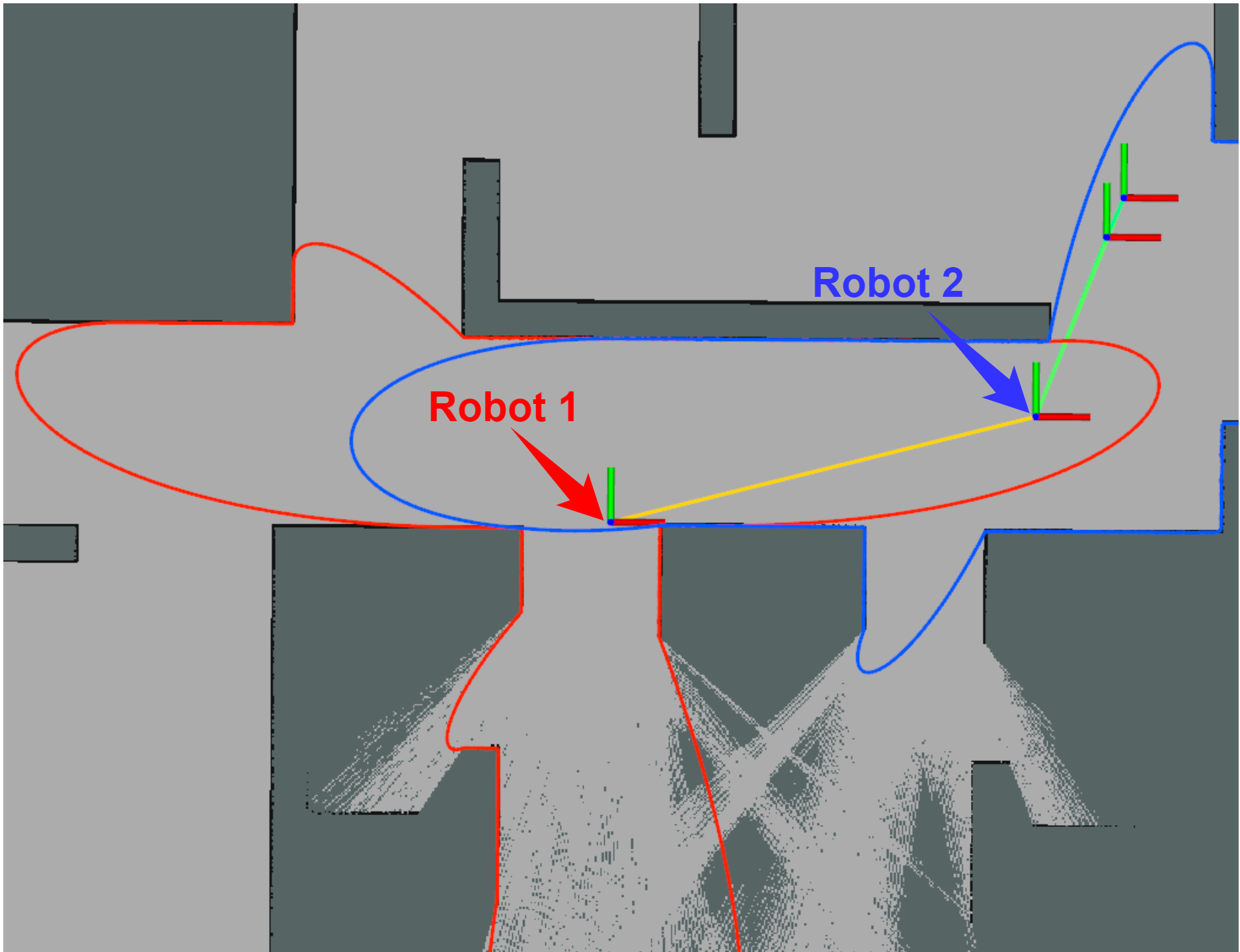}}
\caption{(a) Construction of the visible region~\cite{liu_StarconvexConstrained_2022}; (b) Imbalanced LoS-distance for two robots. 
Robot $1$ is at a higher risk of losing LoS than robot $2$, as robot $1$ is easier to move beyond the visible region (enclosed by the green curve) of robot $2$ after a delta movement.}
\label{fig_visibility_illustration}
\vspace{-10pt}
\end{figure}

As shown in Fig.~\ref{fig_visibility_illustration}(a), given the point cloud measurement $\mathcal{C}_{j}\subseteq \mathbb{R}^{3}$ from the robot $j$, its visible region can be constructed via the following steps~\cite{katz2007direct, liu_StarconvexConstrained_2022, xia_RELINKRealTime_2023}:

1) Augmentation. Add augmented points to fill the gaps in the point cloud $\mathcal{C}_j$;

2) Spherical flipping. Map each point $\q\in \mathcal{C}_j$ along the ray originating from the robot $j$ to a point $\q'$ located outside a sphere of radius $r$, following the equation $\q' = 2r\cdot \frac{\q}{\|\q\|} - \q$. The flipped point cloud is denoted as $\mathcal{C}_{j}'$;

3) Convex hull construction. Generate convex hull of $\mathcal{C}_{j}'$, denoted as $Conv(\mathcal{C}_{j}')$.

4) Inversion. Inverse the convex hull by performing the inverse spherical flipping. 
The inversed shape of the convex hull encloses the visible region of robot $j$.

As shown in Fig.~\ref{fig_visibility_illustration}(a), the visibility of a point $\q_{i}\in \mathbb{R}^{3}$ from the position $\q_{j}\in \mathbb{R}^{3}$ of robot $j$ is determined by the distance $d_{k^*}$, defined as
$d_{k^*} = \max_{k}\{d_k = \n_{k}^{\top}(\q_{i}' - \boldsymbol{a}_k) \vert k = 1, ..., K\}$,
where 
$\n_k$ is the outward normal vector of the $k$-th face of $Conv(\mathcal{C}_{j}')$ composed of $K$ faces;
$\boldsymbol{a}_k\in \mathbb{R}^{3}$ is an arbitrary point on the face $k$; 
$\q_{i}'$ is the flipped point of $\q_{i}$ following the spherical flipping; and the index $k^*$ is defined as $k^{*} = \arg \max_{k\in [1,...,K]} d_{k}$.
If $d_{k^*} > 0$, $\q_i$ is visible from $\q_j$, \ie, they are within line-of-sight with each other; vice verse if $d_{k^*} \le 0$. 
The log-sum-exp relaxation is employed to get a differentiable approximation of $d_{k^*}$, defined as
\begin{equation}
    d_{k^*} \approx \frac{1}{\alpha} \log \left(e^{\alpha d_{1}}+\cdots+e^{\alpha d_{K}}\right),
\label{eq_dx}
\end{equation}
where $\alpha > 0$ is a coefficient that controls the degree of approximation~\cite{liu_StarconvexConstrained_2022}.
The above process is adopted from the hidden point removal (HPR) operator in point cloud visibility analysis~\cite{katz2007direct}, with time complexity of $\mathcal{O}(n\log n)$, where $n$ is the size of a point cloud.

In this paper, we aim to derive the LoS constraints between robots from their point cloud measurements.
Specifically, we first construct the visible regions of robots, and then design potential functions to keep them operating within each other's visible regions to maintain LoS connectivity.

\section{Problem Formulation}

\subsection{Robot Model}

Assume there are a set of robots $\mathcal{R} = \{1, 2, ..., R\}$ to perform navigation tasks in an initially unknown environment.
The position and orientation of a robot $i\in \mathcal{R}$ at time $t$ is denoted as $\q_{i}(t)\in \mathbb{R}^{3}$ and $R_{i}(t)\in \mathcal{SO}(3)$ respectively. We assume the kinematic model of robot $i$ is a single-integrator as follows:
\begin{equation}
    \q_{i}(t+1) = \q_{i}(t) + k^{\text{c}}_{i}\cdot \boldsymbol{u}^{\text{c}}_{i}(t) + k^{\text{n}}_{i}\cdot \boldsymbol{u}^{\text{n}}_{i}(t),
\label{eq_kinematic_model}
\end{equation}
where $\boldsymbol{u}^{\text{c}}_{i}, \boldsymbol{u}^{\text{n}}_{i}\in \mathbb{R}^{3}$ are the velocity commands for connectivity and navigation, respectively;  $k_i^\text{c}, k_i^\text{n}\in \mathbb{R}_{\ge 0}$ are two scaling factors.
Moreover, the combined velocity command is bounded by $\Vert k^{\text{c}}_{i}\cdot \boldsymbol{u}^{\text{c}}_{i}(t) + k^{\text{n}}_{i}\cdot \boldsymbol{u}^{\text{n}}_{i}(t) \Vert_2\le U_{\text{max}}$, where $U_{\text{max}}\in \mathbb{R}_{+}$ is the upper limit of robots' velocities.

\subsection{Connectivity Constraints}
\label{sec_types_of_constraints}

Let $\G(t) = \langle \V, \E(t), \omega \rangle$ be an undirected time-varying graph, where $\V = \{1, ..., R\}$ includes $R$ vertices that correspond to robots in $\mathcal{R}$; $\E(t) \subseteq \V \times \V$ is the edge set at time $t$;
and the weighting function $\omega: \V \times \V \xrightarrow{} \mathbb{R}_{\ge 0}$ evaluates the strength of connectivity between two vertices in the graph.
The relative distance between two robots $i, j\in \mathcal{R}$ is defined as $d_{ij} = d_{ji} = \lVert \q_{i} - \q_{j} \rVert_{2}$.
An edge $(i, j)\in \E$ if and only if the following conditions are satisfied:
\begin{itemize}
    \item (C1) Communication constraints. The relative distance between two robots $i$ and $j$ must be within the communication range $d^{\text{com}}$, \ie, $d_{ij}\le d^{\text{com}}$.
    \item (C2) Line-of-Sight constraints. The two robots must be within each other's line-of-sight, \ie, $\eta \q_{i} + (1-\eta)\q_{j} \notin \mathcal{O}$, $\forall \eta \in [0, 1]$, where $\mathcal{O}\subseteq \mathbb{R}^{3}$ includes obstacles in the environment that are unknown to the robots.
    \item (C3) Collision avoidance constraints. The relative distance between robot $i$ and robot $j$, and their distance to surrounding obstacles must be larger than a safe distance $d^{\text{coll}}$, \ie, $d_{ij}\ge d^{\text{coll}}$.
\end{itemize}

We define a set $\N_{i} = \{j\in \V \vert (i, j)\in \E\}$ as the collection of all neighboring robots of robot $i$ that satisfy above constraints.
The graph $\G$ is referred to as the \emph{connectivity graph} of the robot team. 
Initially, we assume $\G(0)$ is connected.

\subsection{Problem Statement}
\label{eq_problem_statement}

Given a sequence of target points $\mathcal{Z} = \{\boldsymbol{z}^{1}, ..., \boldsymbol{z}^{M}\}\subseteq \mathbb{R}^{3}$ distributed in the free space of an unknown environment, and a group of robots $\mathcal{R} = \{1, ..., R\}$ with kinematic models as in Eq.~(\ref{eq_kinematic_model}).
We assume $M \le R$.
The index of the designated robot to a target $\boldsymbol{z}^m\in \mathcal{Z}$ is denoted as the robot $i_z^m$.
The problem is to find a sequence of velocity commands $\boldsymbol{u}^{\text{c}}_{i}$ and $\boldsymbol{u}^{\text{n}}_{i}$ for each robot $i\in \mathcal{R}$ so that:
(1) there exist a time sequence $0\le t^1, ..., t^M < \infty$ when the corresponding target $\boldsymbol{z}^m$ is visited by the assigned robot $i_z^m$ at time $t^m$, and stay at $\boldsymbol{z}^m$ for a pre-defined time duration $\Delta T$;
(2) the connectivity graph $\G(t)$ is connected for $\forall t \ge 0$.

The unknown obstacles present challenges for LoS maintenance between robots during navigation, which is our main focus in this paper.  
In contrast, the formulations of communication constraints (C1) and inter-robot collision avoidance (part of C3) only depend on robots' relative distances, which can be derived following~\cite{robuffogiordano_PassivitybasedDecentralized_2013}.
The details are provided in the Appendix for completeness.

\section{Methodology}

This section first overviews the potential function design to ensure generalized graph connectivity, and then focuses on the formulation of the LoS constraints from robots' visible regions. 
Finally, we show the connectivity velocity command of each robot can be derived distributively.

\subsection{Potential Function for Fiedler Eigenvalue}

Based on the concept of generalized graph connectivity, we define $\omega(i, j) = A_{ij} = \alpha_{ij}\cdot \beta_{ij} \cdot \gamma_{ij}$ for an edge $(i, j)\in \mathcal{E}$ in the connectivity graph, where $\alpha_{ij}, \beta_{ij}, \gamma_{ij}\in \mathbb{R}_{\ge 0}$ are the weights quantifying the satisfaction of constraints C1, C2, and C3 between two robots $i$ and $j$, respectively.
We define a potential function $V^{\lambda}(\lambda_2) = \frac{1}{\lambda_2 - \lambda_{2}^{\text{min}}}$ to enforce the Fiedler eigenvalue $\lambda_2$ of $\G$ being larger than a preferred lower bound $\lambda_{2}^{\text{min}}>0$ as in~\cite{robuffogiordano_PassivitybasedDecentralized_2013,nestmeyer_DecentralizedSimultaneous_2017}.
Then the velocity command for a robot $i\in \mathcal{R}$ to maintain the graph connectivity can be derived as~\cite{yang_DecentralizedEstimation_2010}
\begin{equation*}
\small
    \boldsymbol{u}^{\text{c}}_i = -\frac{\partial V^{\lambda}(\lambda_2)}{\partial \lambda_2} \cdot \frac{\partial \lambda_2}{\partial \q_{i}} = -\frac{\partial V^{\lambda}(\lambda_2)}{\partial \lambda_2}\cdot \sum_{j\in \mathcal{N}_{i}} \frac{\partial A_{ij}}{\partial \q_i}(v_{2_i} - v_{2_j})^{2},
\label{eq_connectivity_force}
\end{equation*}
where $v_{2_i}$ and $v_{2_j}$ are respectively the $i$-th and $j$-th elements of the normalized eigenvector $\boldsymbol{v}_2$ corresponding to the $\lambda_{2}$. 

In subsequent sections, we will focus on the formulation of $\beta_{ij}(\cdot)$ to quantify the LoS constraints between robots, while the definitions of $\alpha_{ij}(\cdot)$ and $\gamma_{ij}(\cdot)$ are provided in the Appendix due to space limitations.

\subsection{Urgency and Sensitivity Analysis of Losing LoS}

\begin{figure}[t]
\vspace{6pt}
\centering
\subfloat[]{\includegraphics[width=.38\linewidth]{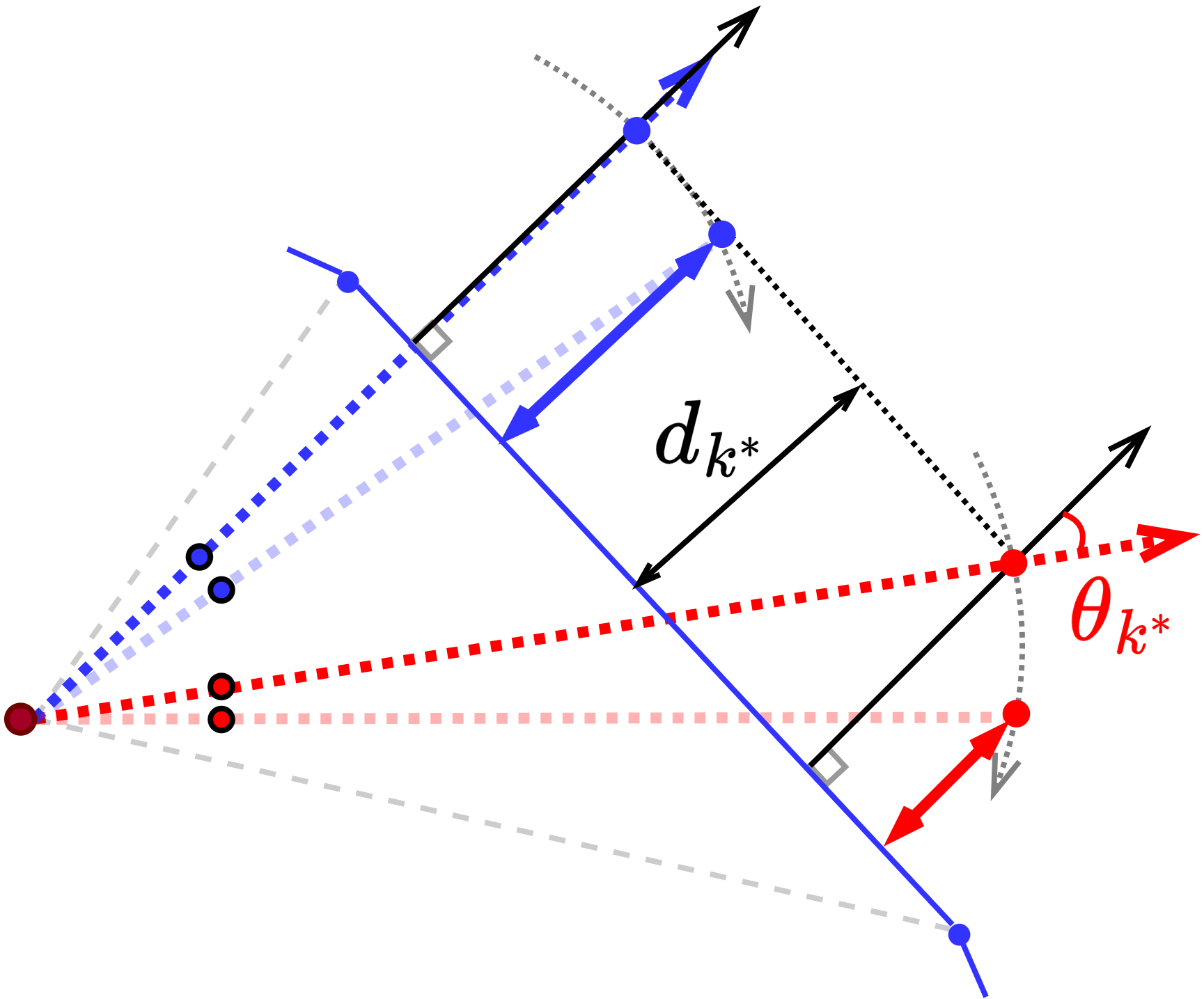}}\hspace{1pt}
\subfloat[]{\includegraphics[width=.6\linewidth]{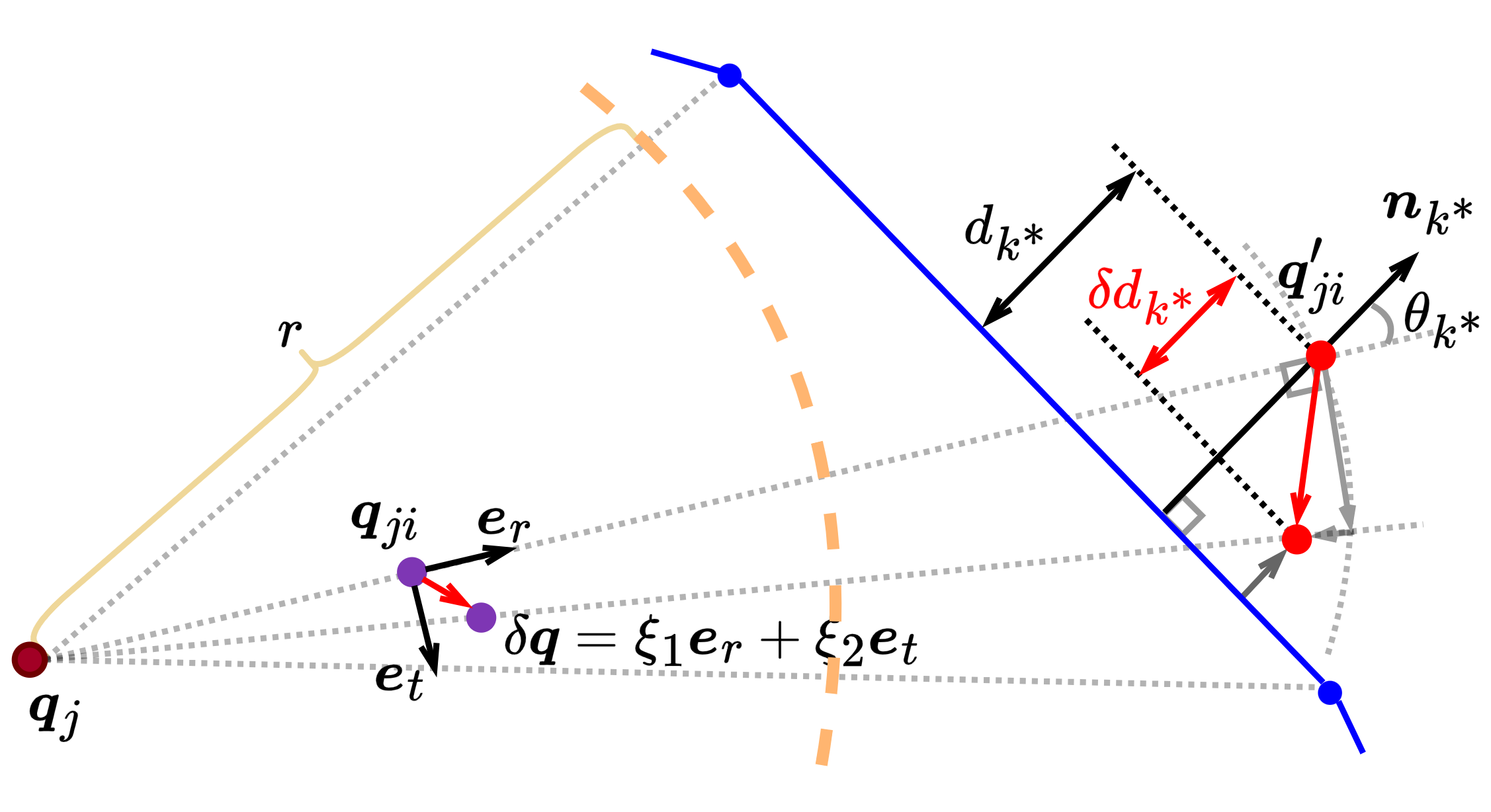}}
\caption{(a) Sensitivity of losing LoS after applying delta movements to two test points (blue and red dots) with a same $d_{k^*}$; (b) Illustration of $\Delta_{d_{k^*}}$ after applying a delta movement $\delta \q$ (red arrow) to $\q_{ji}$ in 2D case.}
\label{fig_sensitivity2}
\end{figure}

This section presents a LoS-distance metric that quantifies both the urgency and sensitivity of losing LoS between two robots. 
While the distance $d_{k^*}$ as in Eq.~(\ref{eq_dx}) can be used to determine the visibility between two robots, it cannot quantify the \emph{sensitivity} of losing LoS as a robot moves. 
The situation is illustrated in Fig.~\ref{fig_sensitivity2}(a), where, although the two cases have the same $d_{k^*}$, the change of $d_{k^*}$ after applying a delta movement to the test point is significantly different.

Here we quantitatively analyze the sensitivity of robot $i$ losing LoS with robot $j$, \ie, moving beyond the visible region of robot $j$.
To begin with, the position of robot $i$ in robot $j$'s local coordinate frame is obtained by $\q_{ji} = R_{j}^{-1}(\q_{i} - \q_{j})\in \mathbb{R}^{3}$.
Recall in Sec.~\ref{sec_preliminary_visibility}, the flipped point of $\q_{ji}$ is calculated as $    \q_{ji}' = 2r\cdot \frac{\q_{ji}}{\lVert \q_{ji} \rVert} - \q_{ji}$.
To quantify the sensitivity of losing LoS, we introduce a disturbance movement $\delta \q$ to $\q_{ji}$, defined as $\delta \q = \xi_{1}\cdot \e_{r} + \xi_{2}\cdot  \e_{t} + \xi_{3}\cdot \e_{n}$,
where $\xi_{1}, \xi_{2}, \xi_{3}\in \mathbb{R}$; $\Vert \delta \q \Vert \le \sigma\in \mathbb{R}_{+}$ with $\sigma\xrightarrow{} 0$; $\e_{r}, \e_{t}, \e_{n}$ are orthogonal basis vectors in radial, tangential, and normal directions.
% where $\xi_{1}, \xi_{2}, \xi_{3}\in [-1, 1]$, $\Vert \delta \q \Vert \le \sigma$ with $\sigma$ being a very small value; and $\e_{r}, \e_{t}, \e_{n}$ are orthogonal basis vectors in radial, tangential, and normal directions.
The notations are illustrated in Fig.~\ref{fig_sensitivity2}(b) in 2D cases. 
Note that $\e_{r}$ is parallel to $\n_{r}$, defined as $\n_{r} = \frac{\q_{ji}'}{\Vert \q_{ji}' \Vert}$.
After applying the delta movement to $\q_{ji}$, the resulting movement of $\q_{ji}'$ is
$\delta\q' \approx  -\xi_{1}\cdot \e_{r} + \frac{2r -\Vert \q_{ji}\Vert}{\Vert \q_{ji}\Vert}(\xi_{2}\cdot  \e_{t} + \xi_{3}\cdot \e_{n})$.
% \begin{equation*}
%     \delta\q' =  -\xi_{1}\cdot \e_{r} + \frac{2r -\Vert \q_{ji}\Vert}{\Vert \q_{ji}\Vert}(\xi_{2}\cdot  \e_{t} + \xi_{3}\cdot \e_{n}).
% \end{equation*}
By projecting to the normal direction $\n_{k^*}$, the resulting change of $d_{k^*}$ is
$ \Delta_{d_{k^*}}
        = 
        (\delta \q')^{\top} \n_{k^*} 
        \approx
        -\xi_{1}\cos{\theta_{k^*}} + \frac{2r -\Vert \q_{ji}\Vert}{\Vert \q_{ji}\Vert}(\xi_{2} + \xi_{3})\sin{\theta_{k^*}}$.
Considering $\xi_{1}, \xi_{2}, \xi_{3}\xrightarrow{} 0$ and that the flipping radius $r$ is typically set to be much larger than $\Vert \q_{ji}\Vert$, we omit the term $-\xi_{1}\cos{\theta_{k^*}}$ and get
\begin{equation*}
\small
    \Delta_{d_{k^*}} \approx \frac{2r -\Vert \q_{ji}\Vert}{\Vert \q_{ji}\Vert}(\xi_{2} + \xi_{3})\sin{\theta_{k^*}} \le \frac{2r -\Vert \q_{ji}\Vert}{\Vert \q_{ji}\Vert}\cdot \sqrt{2}\sigma\cdot \sin{\theta_{k^*}}.
\label{eq_approximate}
\end{equation*}
According to the above expression, in the worst case, the sensitivity of $d_{k^*}$ w.r.t. the robot's delta movement is determined by $\sin{\theta_{k^*}}$, \ie, a larger $\vert \theta_{k^*} \vert$ will introduce larger potential decrease of $d_{k^*}$, and vice versa. 
The analysis is aligned with the illustrations in Fig.~\ref{fig_sensitivity2}(a).

Therefore, we define a sensitivity-encoded LoS-distance metric that quantifies the distance from robot $i$ to the visible region of robot $j$ as 
$\hat{d}^{\text{los}}_{ji} = d_{k^{*}}\cdot \cos{\theta_{k^{*}}}$.
A differentiable approximation of this metric is defined as 
\begin{equation}
\small
   \hat{d}^{\text{los}}_{ji} 
   = 
   \frac{1}{\alpha} \log(\sum_{k = 1}^{K}e^{\alpha d_{k}}) 
   \cdot 
   \frac{\sum_{k = 1}^{K}e^{\alpha d_{k}}\cdot \cos{\theta_{k}}}{\sum_{k = 1}^{K}e^{\alpha d_{k}}},
\label{eq_final_distance_function}
\end{equation}
where $\cos{\theta_{k}} = \n_{r}^{\top} \n_{k}$.
The rationale behind this metric is, to increase $\hat{d}^{\text{los}}_{ji}$, the movement of robot $i$ should not only contribute to a larger $d_{k^{*}}$ (to reduce urgency), but also a larger $\cos{\theta_{k^{*}}}$ to reduce the sensitivity of losing LoS with robot $j$.
The derivative of $\hat{d}^{\text{los}}_{ji}$ w.r.t. $\q_{ji}'$ is calculated as
\begin{equation}
\small
\begin{aligned}
\frac{\partial \hat{d}^{\text{los}}_{ji}}{\partial \q_{ji}'} 
=& 
\cos{\hat{\theta}_{k^{*}}}
\cdot \hat{\n}_{k^{*}}^{\top} + 
\frac{\hat{d}_{k^{*}}}{\Vert \q_{ji}' \Vert}(\hat{\n}_{k^{*}}^{\top} - \cos{\hat{\theta}_{k^{*}}}\cdot \n_{r}^{\top})\\
&+ \alpha\cdot \hat{d}_{k^{*}} (
\frac{\widehat{\cos{\theta_{k^{*}}}
\cdot \n_{k^{*}}^{\top}}}{\cos{\hat{\theta}_{k^{*}}}}
 - \hat{\n}_{k^{*}}^{\top}) 
,
\end{aligned}
\label{eq_original_derivative}
\end{equation}
where 
$\cos{\hat{\theta}_{k^{*}}} = \frac{\sum e^{\alpha d_{k}}\cdot \cos{\theta_{k}}}{\sum e^{\alpha d_{k}}}$;
$\hat{\n}_{k^{*}}^{\top} = \frac{\sum e^{\alpha d_{k}}\cdot \n_{k}^{\top}}{\sum e^{\alpha d_{k}}} $;
$\hat{d}_{k^{*}} = \frac{\hat{d}^{\text{los}}_{ji}}{\cos{\hat{\theta}_{k^{*}}}}$;
$\widehat{\cos{\theta_{k^{*}}}
\cdot \n_{k^{*}}^{\top}}
=
\frac{\sum e^{\alpha d_{k}} (\cos{\theta_{k}}\cdot \n_{k}^{\top})}{\sum e^{\alpha d_{k}}}$.
Assuming ideal log-sum-exp approximation, Eq.~(\ref{eq_original_derivative}) can be simplified as
\begin{equation}
        \frac{\partial \hat{d}^{\text{los}}_{ji}}{\partial \q_{ji}'} 
        \approx
        \cos{\theta_{k^{*}}}\cdot \n_{k^{*}}^{\top} + \frac{d_{k^{*}}}{\Vert \q_{ji}' \Vert}(\n_{k^{*}}^{\top} - \cos{\theta_{k^{*}}}\cdot \n_{r}^{\top}).
\label{eq_gradient_sensitivity}
\end{equation}
Note that when $\theta_{k^{*}} = 0$, \ie, $\n_{k^{*}} = \n_{r}$, the derivative is calculated as $    \frac{\partial \hat{d}^{\text{los}}_{ji}}{\partial \q_{ji}'} \Big|_{\theta_{k^{*}}=0}
    \approx 
    \n_{k^{*}}^{\top}$,
which is the same as the derivative when using Eq.~(\ref{eq_dx}) to measure the LoS-distance.

\begin{remark}
    The new LoS-distance metric has two benefits: (1) when $d_{k^*}\ge 0$, it holds that $\hat{d}^{\text{los}}_{ji} \le d_{k^*}$, which means that $\hat{d}^{\text{los}}_{ji}$ will incur a larger connectivity velocity in Eq.~(\ref{eq_def_beta}) when robots are more sensitive to losing LoS with a same $d_{k^*}$;
    (2) compared with using $d_{k^*}$ alone as in Eq.~(\ref{eq_dx}), the derivative in Eq.~(\ref{eq_gradient_sensitivity}) has an additional component (\ie, the second term on the right-hand side) that decreases $\vert \theta_{k^*} \vert$, hence reduces sensitivity of losing LoS.
\end{remark}

Finally, the derivative of $\hat{d}^{\text{los}}_{ji}$ w.r.t. $\q_{ji}$ is calculated as
\begin{equation*}
    \frac{\partial \hat{d}^{\text{los}}_{ji}}{\partial \q_{ji}} 
    =
    \frac{\partial \hat{d}^{\text{los}}_{ji}}{\partial \q_{ji}'} \cdot \frac{\partial \q_{ji}'}{\q_{ji}}
    =
    \frac{\partial \hat{d}^{\text{los}}_{ji}}{\partial \q_{ji}'} (\frac{2r}{\|\q_{ji}\|}-\frac{2r}{\|\q_{ji}\|^{3}}\cdot \q_{ji} \q_{ji}^{\top}-1).
\end{equation*}

\subsection{Fusion of Imbalanced LoS-Distance}

This section proposes the potential function $\beta(\cdot)$ that encodes the LoS-distances between the robots $i$ and $j$ into the weight of the edge $(i, j)\in \mathcal{E}$. 
To ensure that they are within each other's LoS, it is required that $\hdlos_{ij}, \hdlos_{ji} \ge 0$.

However, directly designing a potential function for $\hdlos_{ij}$ and $\hdlos_{ji}$ is not applicable, because it is often the case that $\hdlos_{ij}\ne \hdlos_{ji}$, which will finally lead to asymmetric values of the edge weights $A_{ij}$ and $A_{ji}$.
In fact, the different values of $\hdlos_{ij}$ and $\hdlos_{ji}$ reflect the imbalanced urgency of losing LoS for two robots, as illustrated in Fig.~\ref{fig_visibility_illustration}(b).
To keep the Laplacian matrix symmetric, we need to find a differentiable function to fuse $\hdlos_{ij}$ and $\hdlos_{ji}$ while reflecting the urgency of losing LoS between the two robots.
As such urgency is determined by the smaller value between $\hdlos_{ij}$ and $\hdlos_{ji}$,
an intuitive option is to use a differentiable approximation of the $\min(\cdot)$ function such as $\text{Softmin}(\cdot)$ function to take the smaller value.
However, this will lead to poor cooperation between robots to maintain LoS, as verified in Sec.~\ref{sec_fusion_evaluation}.

To address this issue, we define a LoS-distance fusion function $d^{\text{los}}(\cdot)$ as
\begin{equation}
\small
\dlos_{ij} = \frac{\hat{d}^{\text{los}}_{ij}+c}{\hat{d}^{\text{los}}_{ij} + \hat{d}^{\text{los}}_{ji}+2c}\cdot \hat{d}^{\text{los}}_{ji} + \frac{\hat{d}^{\text{los}}_{ji}+c}{\hat{d}^{\text{los}}_{ij} + \hat{d}^{\text{los}}_{ji}+2c}\cdot \hat{d}^{\text{los}}_{ij}
\label{eq_weighted_average}
\end{equation}
when $\hdlos_{ij}, \hdlos_{ji} \ge \dlos_{\text{min}}$.
The Eq.~(\ref{eq_weighted_average}) is essentially a weighted sum of $\hdlos_{ij}$ and $ \hdlos_{ji}$,
where $c\in \mathbb{R}_{\ge0}$ is a constant parameter that controls whether $\dlos_{ij}$ depends more on the smaller value among $\hdlos_{ij}$ and $ \hdlos_{ji}$ (when $c$ takes a small value) or their averaged value (otherwise). 
% In Eq.~(\ref{eq_weighted_average}), $\dlos_{ij}$ is mainly determined by the smaller value between $\hat{d}^{\text{los}}_{ij}$ and $\hat{d}^{\text{los}}_{ji}$, which quantify the urgency of losing LoS between the two robots.
Particularly, the gradient of $\dlos_{ij}$ w.r.t. $\hat{d}^{\text{los}}_{ij}$ and $\hat{d}^{\text{los}}_{ji}$ is less imbalanced that using the $\text{Softmin}(\cdot)$ function, and can be regulated by the parameter $c$, which facilitates the less urgent robot's movement to maintain LoS with its more urgent neighbor.

According to Eq.~(\ref{eq_weighted_average}), it holds that $\dlos_{ij} = \dlos_{ji}$.
Therefore, we design the potential function for $\dlos_{ij}$ or $\dlos_{ji}$ that naturally ensures $\beta_{ij} = \beta_{ji}$ as follows:
\begin{equation}  
\small
\beta_{ij} =\left \{
    \begin{aligned}
        &0, &0\le d^{\text{los}}_{ji} < d^{\text{los}}_{\text{min}}\\
        &\frac{k_s}{2}[1-\cos(\frac{d^{\text{los}}_{ji} - d^{\text{los}}_{\text{min}}}{d^{\text{los}}_{\text{max}} - d^{\text{los}}_{\text{min}}})\pi], &d^{\text{los}}_{\text{min}}\le d^{\text{los}}_{ji} < d^{\text{los}}_{\text{max}},\\
        &k_s, &d^{\text{los}}_{ji} \ge d^{\text{los}}_{\text{max}},
    \end{aligned}
    \right.
\label{eq_def_beta}
\end{equation}
where $d^{\text{los}}_{\text{max}} > 0$ is the distance from which the LoS between two robots start being affected, until they lose LoS when $d^{\text{los}}_{ji} < d^{\text{los}}_{\text{min}}$.
Finally, the derivative of $\beta_{ij}$ w.r.t. $\q_{ji}$ is obtained as $\frac{\partial \beta_{ij}}{\partial \q_{ji}} = \frac{\partial \beta_{ij}}{\partial \dlos_{ji}} \cdot \frac{\partial \dlos_{ji}}{\partial \hat{d}^{\text{los}}_{ji}} \cdot \frac{\partial \hat{d}^{\text{los}}_{ji}}{\partial \q_{ji}}$.
As $\q_{ji}$ is defined in robot $j$'s local frame, the gradient is transformed to the world frame as $\frac{\partial \beta_{ij}}{\partial \q_{i}} = \frac{\partial \beta_{ij}}{\partial \q_{ji}}
    \cdot 
    \frac{\partial \q_{ji}}{\partial\q_{i}}
    = \frac{\partial \beta_{ij}}{\partial \q_{ji}}
    \cdot 
    R_{j}^{-1}$.

\subsection{Derivation of Connectivity Velocity}

After defining $\beta_{ij}(\cdot)$ as in Eq.~(\ref{eq_def_beta}), and $\alpha_{ij}(\cdot)$ and $\gamma_{ij}(\cdot)$ as in the Appendix, the connectivity velocity of each robot $i\in \mathcal{R}$ can be calculated as
\begin{equation}
\small
      \boldsymbol{u}^{\text{c}}_i =  -\frac{1}{(\lambda_2 - \lambda_{2}^{\text{min})^2}}\cdot \sum_{j\in \mathcal{N}_{i}} \frac{\partial A_{ij}}{\partial \q_{i}} \cdot (v_{2_i} - v_{2_j})^{2},
\label{eq_connect_force_final}
\end{equation}
where $\frac{\partial A_{ij}}{\partial \q_{i}} = \frac{\partial\alpha_{ij}}{\partial \q_{i}}\cdot \gamma_{ij}\cdot \beta_{ij}  + \alpha_{ij}\cdot \frac{\partial \gamma_{ij}}{\partial \q_{i}}\cdot \beta_{ij} + \alpha_{ij}\cdot \gamma_{ij}\cdot \frac{\partial \beta_{ij}}{\partial \q_i}$.

\begin{proposition}
    The connectivity velocity in Eq.~(\ref{eq_connect_force_final}) can be calculated distributively by each robot, requiring only one-hop communication with its neighbors (See proof in the Appendix).
\end{proposition}

\section{LoS-Constrained Navigation Framework}

\begin{figure}[t]
\vspace{6pt}
\centering\includegraphics[width=\linewidth]{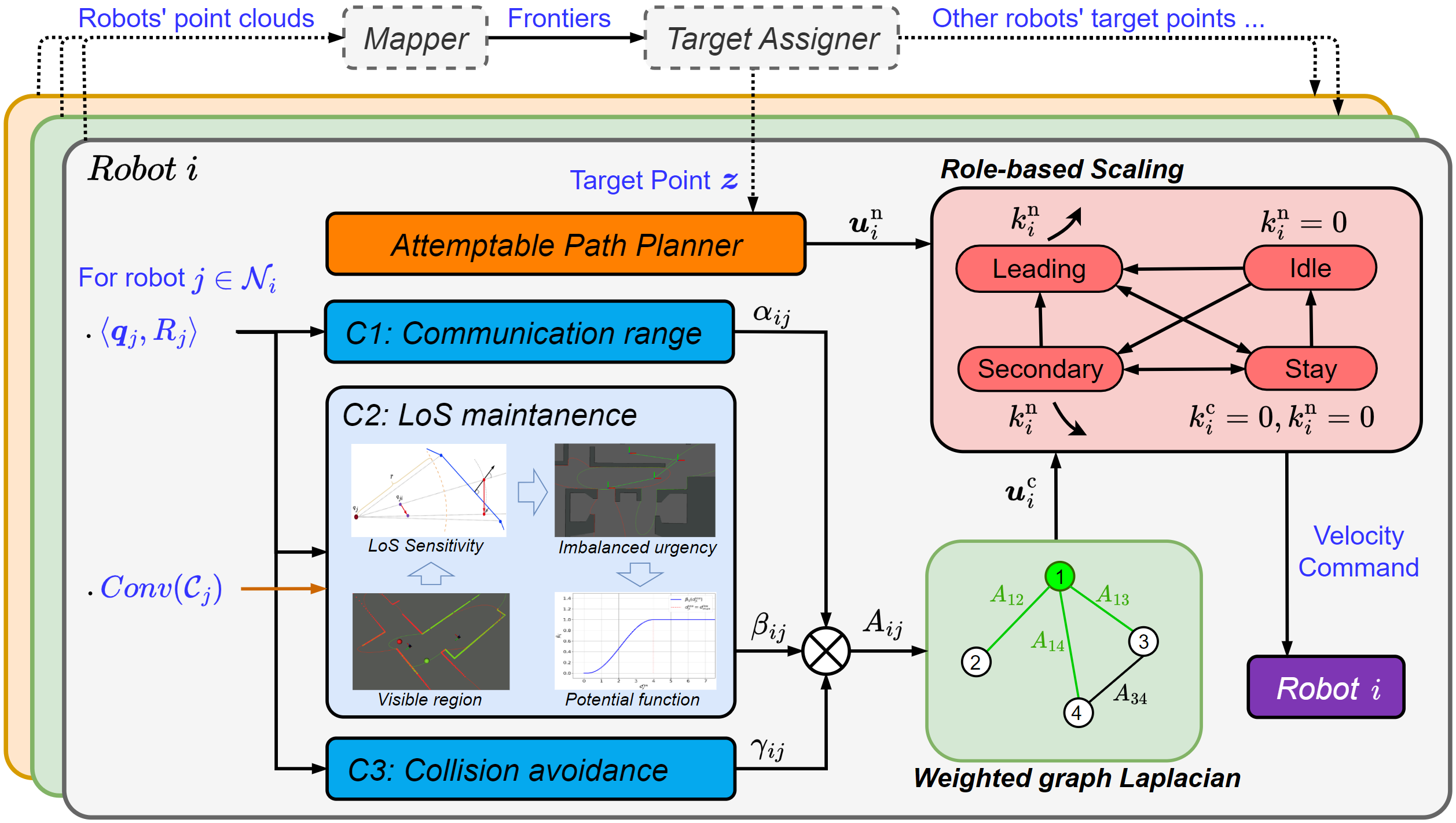}
\vspace{-23pt} 
\caption{The framework of LoS-constrained multi-robot exploration.}
 \label{fig_framework}
 \vspace{-5pt}
\end{figure}

This section introduces how to derive the navigation velocity for robots to perform the navigation tasks as specified in Sec.~\ref{eq_problem_statement}.
To achieve mapless navigation, we employ an attemptable path planner, called FAR Planner~\cite{yang_FARPlanner_2022}, which attempts to find a reference path to the goal point while updating a visibility graph online for quick replanning.
% \footnote{Note that our method does not rely on any specific implementation of path planners, as long as it allows attemptable path planning in unknown environments.}. 
Given a target $\boldsymbol{z}^{m}$ for a robot $i\in \mathcal{R}$, the FAR Planner will serve intermediate waypoints that are visible to robot $i$ step by step toward $\boldsymbol{z}^{m}$.
Assume the waypoint served by the FAR Planner at time $t$ is $\p_{i}(t)\in \mathbb{R}^3$, the navigation velocity is calculated as
$\boldsymbol{u}^{\text{n}}_{i}(t) = \frac{\p_{i}(t) - \q_{i}(t)}{\Vert \p_{i}(t) - \q_{i}(t) \Vert}$.

Moreover, considering the potential conflicts between robots' navigation directions, we adopted the role-based scaling mechanism proposed in~\cite{amigoni_MultirobotExploration_2017} to improve their navigation efficiency.
The robot that is closest to its target point will be elected as the \emph{leading} robot, who will be assigned a larger scaling factor $k_{i}^{\text{n}}$ than the \emph{secondary} robots, 
hence dominating the others' behavior when they have conflicts.
The overall LoS-constrained multi-robot navigation framework and details about role-based scaling mechanism are shown in Fig.~\ref{fig_framework}, and evaluated in Sec.~\ref{sec_experiment}.

\begin{proposition}
    The connectivity of the graph $\G$ is always guaranteed if the navigation velocity $\boldsymbol{u}^{\text{n}}_{i}$ is bounded (See proof in the Appendix).
\label{prop_stability}
\end{proposition}

\section{Experiments and Analysis}
\label{sec_experiment}

\begin{figure}[t]
\centering\includegraphics[width=\linewidth]{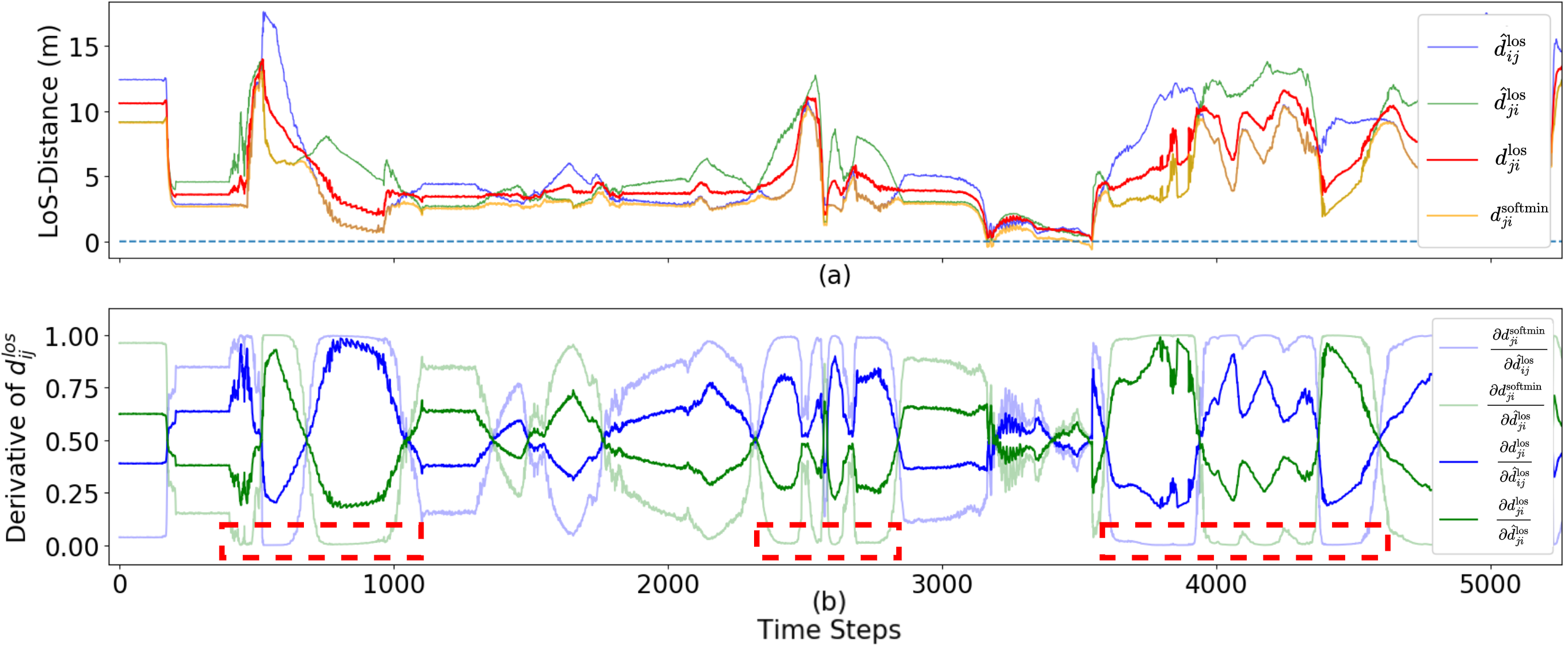}
\vspace{-22pt} 
\caption{(a) Comparison of the fused LoS-distances between two robots using Eq.~(\ref{eq_weighted_average}) and using the $\text{Softmin}(\cdot)$ function;
(b) Comparison of the derivatives of Eq.~(\ref{eq_weighted_average}) (solid lines) and that of the $\text{Softmin}(\cdot)$ function (transparent lines).}
 \label{fig_compare_distance_functions}
\vspace{-15pt}
\end{figure}

This section first compares the effectiveness of different LoS-distance fusion methods on handling imbalanced LoS-distance between robots.
For connectivity maintenance, as most existing methods assume known environments, it is difficult to conduct fair comparisons with them.
Instead, we verify the effectiveness of our method in LoS-constrained exploration of complex unknown environments as shown in Fig.~\ref{fig_mapping_results}, and provide the quantitative analysis of the proposed method.
The experiments are conducted in Gazebo on a desktop with an i9-13900 CPU and 32 GB of RAM, where robots are equipped with 2D LiDARs with 360 degrees of FoV.
The laser points that hit neighboring robots are removed before constructing the visible regions.
The multi-robot mapping algorithm we used is adopted from Karto~\cite{Konolige_Karto_2010}.

\subsection{Implementation Details}
\label{sec_implementatin_details}

\textbf{Obstacle Avoidance}: While the high-level path planner handles obstacle avoidance, we also design potential functions for obstacle avoidance to ensure safety. 
Specifically, each robot $i\in \mathcal{R}$ will select the closest laser point $\q \in \mathcal{C}_{i}$ as its nearest obstacle point, which will be treated as a static \emph{virtual} neighboring robot. Then the potential function for inter-robot collision avoidance as in~\cite{robuffogiordano_PassivitybasedDecentralized_2013} can be directly applied. 
The details are provided in the Appendix.

\textbf{Target Generation}:
During multi-robot exploration, a sequence of targets $\mathcal{Z}$ is generated as the frontiers in the environment, which are located at boundaries between known and unknown areas.
The frontiers are assigned to robots based on their distance to robots and the information gain.

\subsection{Evaluation of LoS-Distance Fusion Function}
\label{sec_fusion_evaluation}

This section compares the proposed fusion function in Eq.~(\ref{eq_weighted_average}) with the $\text{Softmin}(\cdot)$ function, which is a differentiable relaxation of the $\min(\cdot)$ function, defined as 
$d^{\text{softmin}}_{ij}= -\frac{1}{\beta}\log(e^{-\beta\cdot \hat{d}^{\text{los}}_{ij}} + e^{-\beta\cdot \hat{d}^{\text{los}}_{ji}})$,
where $\beta>0$ is a coefficient that controls the degree of approximation.
We show in Fig.~\ref{fig_compare_distance_functions}(a) the LoS-distance between two robots during the exploration of Env.~2 in Fig.~\ref{fig_mapping_results}.
The $\text{Softmin}(\cdot)$ provides a well lower-bound approximation of the smaller value between $\hat{d}^{\text{los}}_{ij}$ and $\hat{d}^{\text{los}}_{ji}$. 
However, when the difference between $\hat{d}^{\text{los}}_{ij}$ and $\hat{d}^{\text{los}}_{ji}$ is significant, the derivative corresponding to the larger value (\ie, the less urgent robot) is close to zero, as shown in Fig.~\ref{fig_compare_distance_functions}(b).
In that case, one robot may lose LoS with another robot because the less urgent robot (with a derivative close to zero) will not move to help maintain LoS. 
In contrast, the proposed fusion function $\dlos(\cdot)$ not only captures the LoS-distance of the more urgent robot, but also generates the less imbalanced gradients for both robots so that they will driven by the connectivity velocity to maintain LoS collaboratively.

\subsection{Evaluation of Connectivity in Multi-Robot Navigation}

\begin{figure}[!t]
\centering
\subfloat[Env.~1 ($87m\times 69m$)]{\includegraphics[width=.562\linewidth]{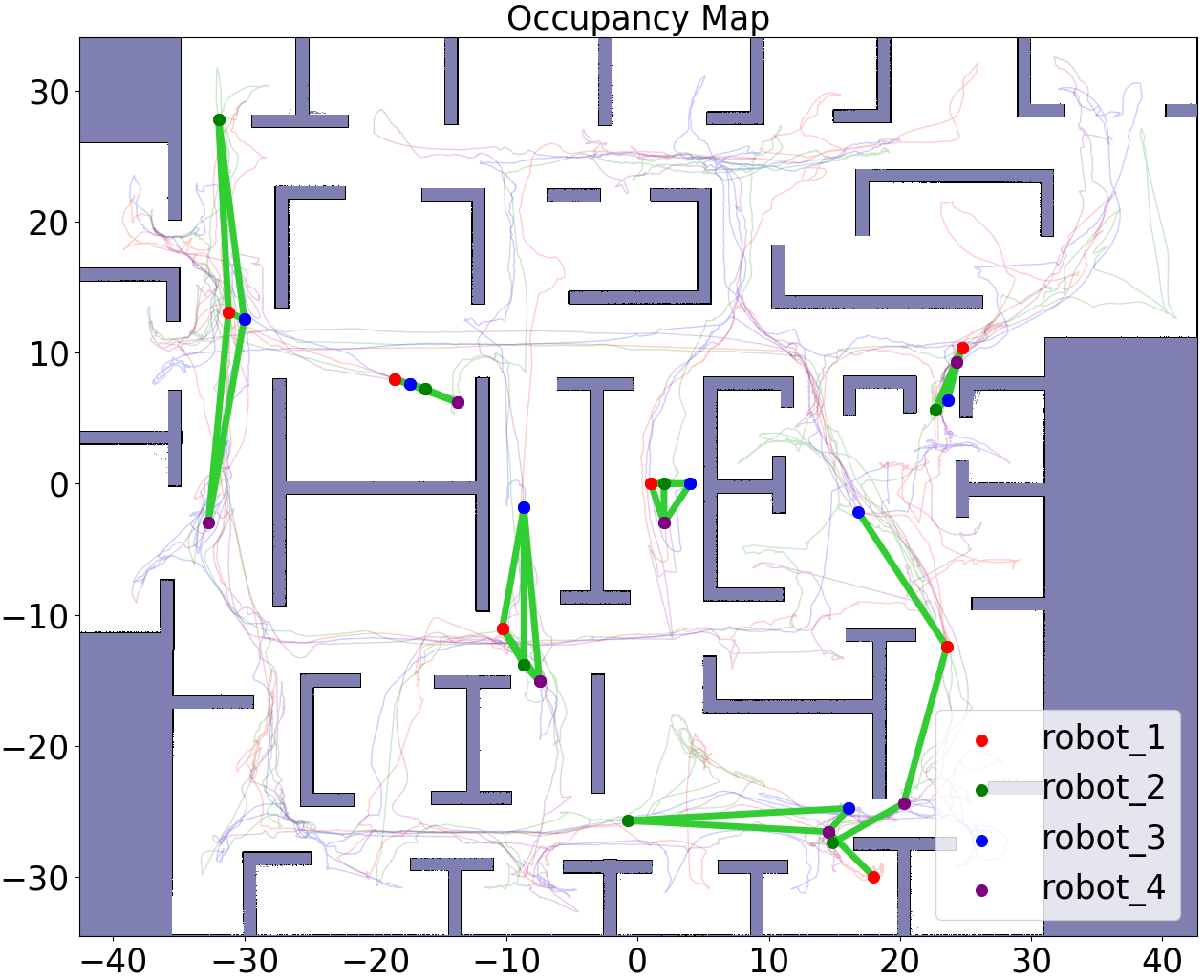}}\hspace{1pt}
\subfloat[Env.~2 ($40m\times 57m$)]{\includegraphics[width=.418\linewidth]{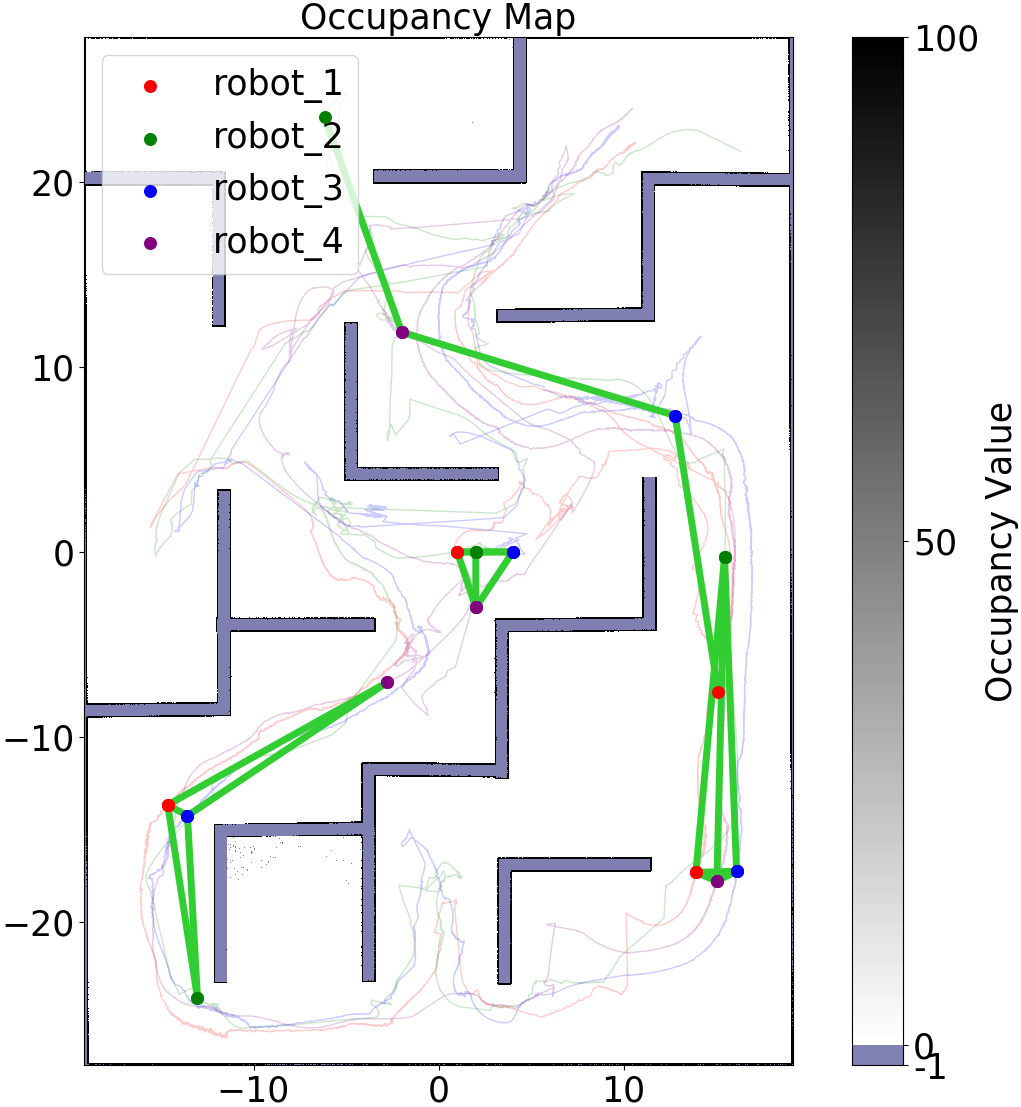}}\\
\subfloat[Real-world demonstration]{\includegraphics[width=.8\linewidth]{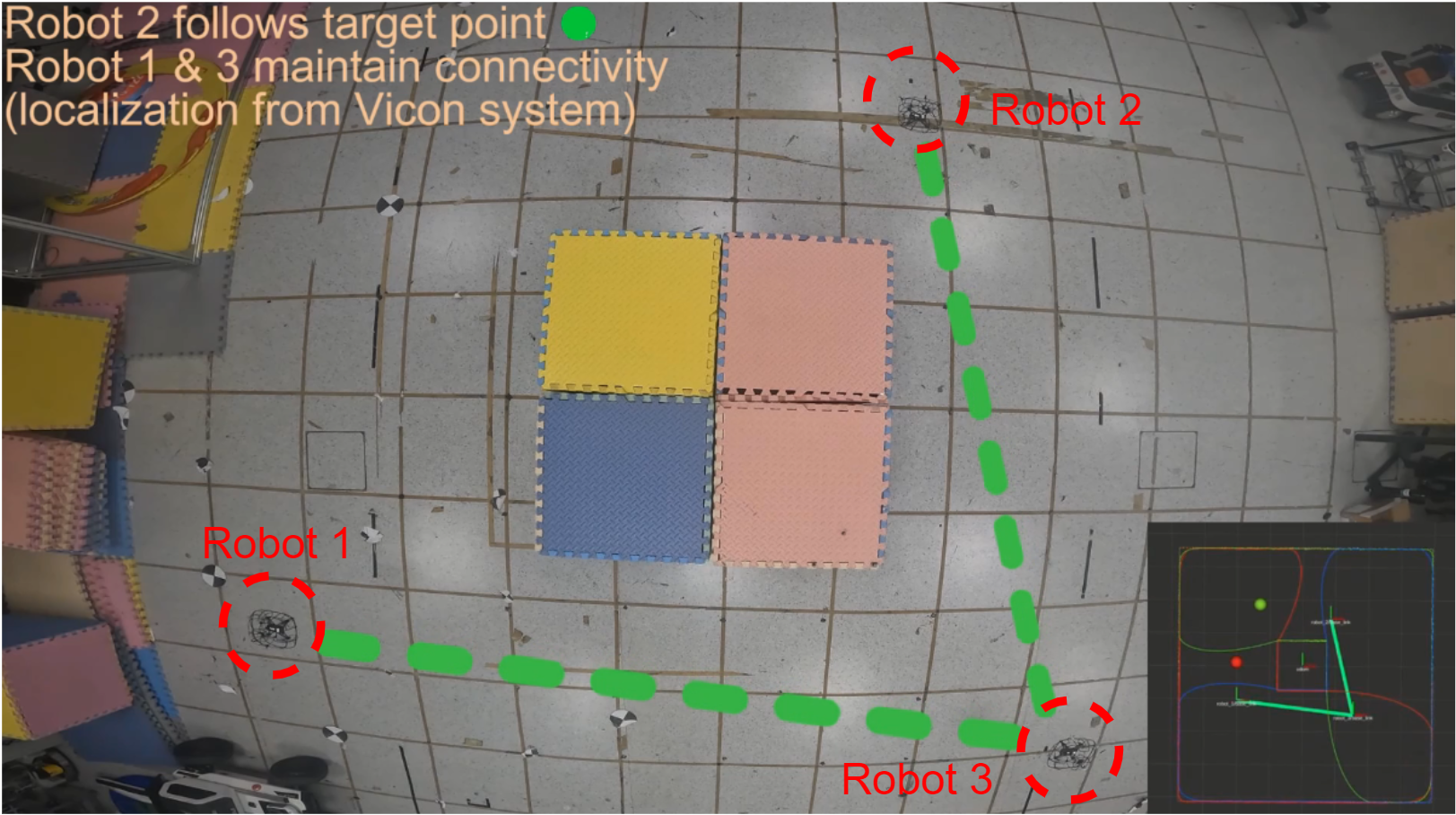}}
\caption{Snapshots of robots' connectivity graphs (green edges) during the exploration of the Env.~1 (a) and Env.~2 (b), and during a navigation task in a real-world demonstration (c). In (a) and (b), the occupancy maps are incrementally built from multi-robot collaborative mapping. Note that the topology of robots' connectivity graph varies over time.}
\label{fig_mapping_results}
\vspace{-10pt}
\end{figure}

\begin{figure}[t]
\centering\includegraphics[width=\linewidth]{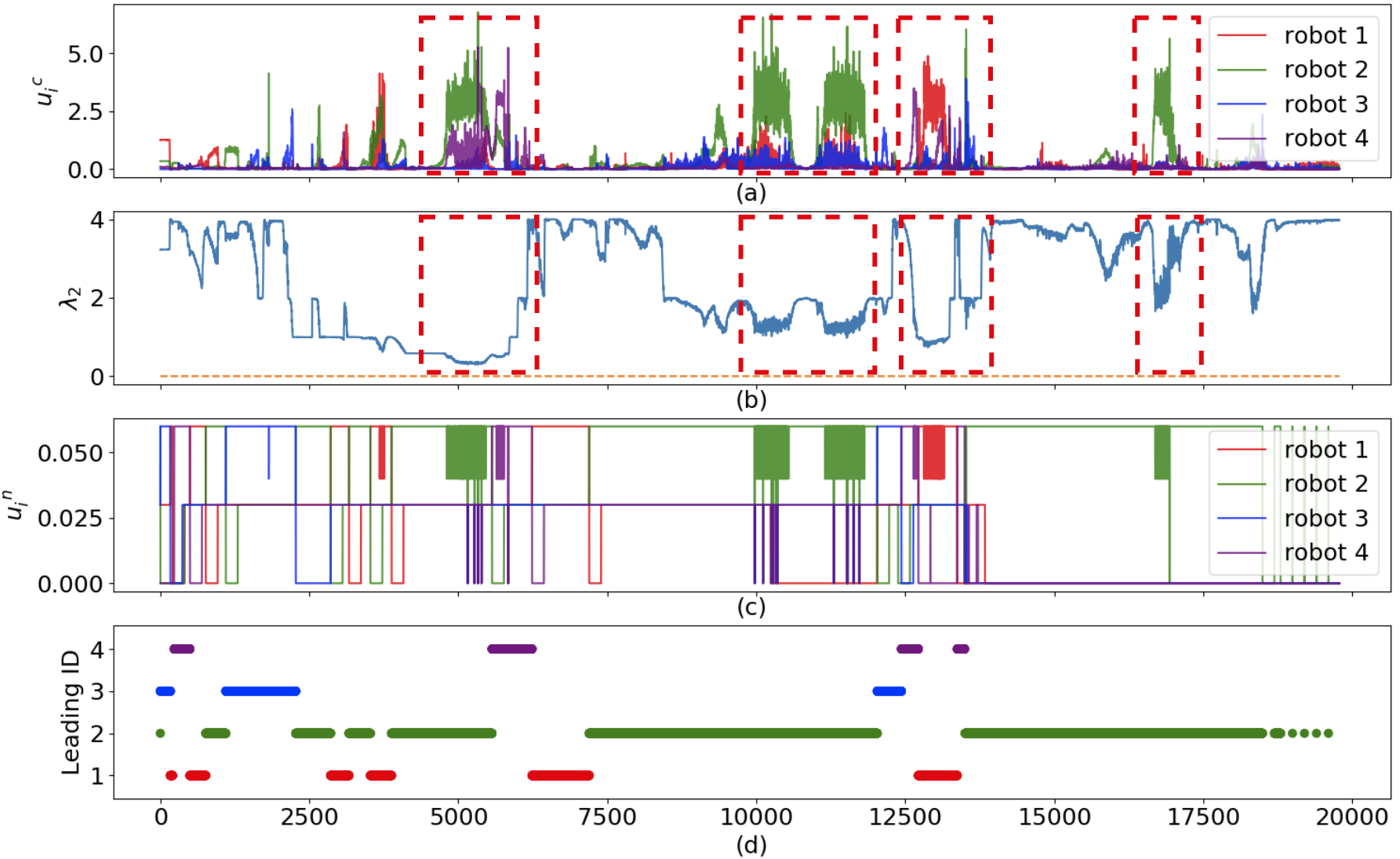}
\vspace{-20pt} % 调整这个数值来改变距离
\caption{Record of parameters during multi-robot exploration of Env.~$2$.
The shared x-axis represents time steps.
(a) the magnitude of connectivity velocity $\boldsymbol{u}^{\text{c}}_{i}$;
(b) the fluctuation of $\lambda_2$ of the connectivity graph $\G$;
(c) the magnitude of navigation velocity $\boldsymbol{u}^{\text{n}}_{i}$ after role-based scaling;
(d) ID of the leading robot over time, which corresponds to a larger $\boldsymbol{u}^{\text{n}}_{i}$ in (c).}
 \label{fig_connect_force}
\vspace{-10pt}
\end{figure}

We evaluate the multi-robot connectivity maintenance of the proposed methods during the exploration of complex unknown environments, as shown in Fig.~\ref{fig_mapping_results}.
The environments are filled with sharp turns and dense walls, making it challenging for robots to maintain LoS.
Despite this, our method successfully maintains LoS-connectivity between robots throughout the exploration, as verified by the positive Fiedler eigenvalue $\lambda_2$ in Fig.~\ref{fig_connect_force}(b). 
The satisfaction of constraints (C1, C2, C3) is also illustrated by the robots' exploration trajectories and their connectivity graphs over time in Fig.~\ref{fig_mapping_results}.
Furthermore, we quantitively compare the connectivity velocity $\boldsymbol{u}^{\text{c}}_{i}$ and navigation velocity $\boldsymbol{u}^{\text{n}}_{i}$ under varying Fiedler eigenvalue $\lambda_2$.
As shown in Fig.~\ref{fig_connect_force}(a) and (b), when $\lambda_2$ approaches zero, the connectivity velocity grows unbounded while the navigation velocity $\boldsymbol{u}_{i}^{\text{n}}$ is bounded as in Fig.~\ref{fig_connect_force}(c), hence the combined velocity commands are dominated to maintain the connectivity.
The role-based scaling of the navigation velocity is shown in Fig.~\ref{fig_connect_force}~(c) and (d), where the leading robot has a larger magnitude of the navigation velocity to resolve deadlocks when two robots move in opposite directions.

We also demonstrate the proposed method in the real-world, as shown in Fig.~\ref{fig_mapping_results}(c), where three DJI Tello Mini Drones with simulated 2D laser with 360 degrees of FoV are deployed. One drone follows a sequence of user-specified goals while the others move to maintain LoS connectivity.
The results show that the robots can always maintain LoS connectivity when navigating in an unknown environment with obstacles.
More experiments and details can be found in the attached video.

\subsection{Discussion}

\textbf{Multi-sensor fusion:} 
Our method accepts different LiDAR configurations, where multiple point cloud measurements can be fused after calibration into a single, larger point cloud to directly determine the visible region for a robot with a larger field of view, without additional adaptation.

\textbf{Kinematic model of robots:} 
Although we assume a first-order kinematic model as in Eq.~(\ref{eq_kinematic_model}), the velocity commands $\boldsymbol{u}^{\text{c}}_{i}$ and $\boldsymbol{u}^{\text{n}}_{i}$ can be mapped to control mobile robots with unicycle models using near-identity diffeomorphism as in~\cite{robotarium_CSM_2020}.

\section{CONCLUSIONS}

This paper presents a LoS-constrained multi-robot navigation method that ensures connectivity among robots during exploration in unknown environments. 
We eliminate assumptions on known environment models by deriving LoS constraints directly from robots' real-time point cloud measurements, by adopting techniques in point cloud visibility analysis.
The effectiveness of the proposed method for connectivity maintenance has been verified in experiments.
% can always ensure the connectivity of robots while exploring unknown environments, allowing flexible time-varying connectivity topology.
The proposed method can be applied to explore environments with risks and complications, where mutual observations between robots are critical for situational awareness.
Future work is to study methods that allow temporary disconnection (caused by dynamic or small obstacles) to further improve robots' navigation efficiency under connectivity constraints.


%%%%%%%%%%%%%%%%%%%%%%%%%%%%%%%%%%%%%%%%%%%%%%%%%%%%%%%%%%%%%%%%%%%%%%%%%%%%%%%%

%%%%%%%%%%%%%%%%%%%%%%%%%%%%%%%%%%%%%%%%%%%%%%%%%%%%%%%%%%%%%%%%%%%%%%%%%%%%%%%%

%%%%%%%%%%%%%%%%%%%%%%%%%%%%%%%%%%%%%%%%%%%%%%%%%%%%%%%%%%%%%%%%%%%%%%%%%%%%%%%%
\bibliographystyle{ieeetr} %ieeetr国际电气电子工程师协会期刊
\bibliography{main} % ref就是之前建立的ref.bib文件的前缀

\newpage
\section*{Appendix}

\subsection{Formulation of Communication and Collision Avoidance Constraints}

 Note we adopted the formulations in~\cite{robuffogiordano_PassivitybasedDecentralized_2013} to derive the communication radius (C1) and collision avoidance constraints (C3) in this work.

\subsubsection{C1: Communication Radius Constraints $\alpha_{ij}(\cdot)$}
\label{sec_communication_radius}

For two robots $i, j\in \mathcal{R}$, their relative distance is calculated as $d_{ij} = \| \q_{i} - \q_{j} \|$.
The potential function quantifies the communication constraints between the two robots is defined as
\begin{equation}
\alpha_{ij}=\left\{
\begin{aligned}
&k_{\alpha}, &0 \le d_{ij} \le d^{\text{com}}_{\text{min}}\\
& \frac{k_{\alpha}}{2}[1 + \cos(\frac{d_{ij} - d^{\text{com}}_{\text{min}}}{d^{\text{com}}_{\text{max}} - d^{\text{com}}_{\text{min}}})\pi ], &d^{\text{com}}_{\text{min}} < d_{ij} \le d^{\text{com}}_{\text{max}} \\
&0, &d_{ij} > d^{\text{com}}_{\text{max}}
\end{aligned}
\right.
\label{eq_gamma_ij}
\end{equation}
where $d^{\text{com}}_{\text{min}} \ge 0$ is the distance at which communication reliability starts to decrease; and the communication breaks when $d_{ij} > d^{\text{com}}_{\text{max}}$.
The derivative of $\alpha_{ij}$ w.r.t. $\q_{i}$ is calculated as
\begin{equation}
    \frac{\partial \alpha_{ij}}{\partial \q_i} = \frac{\partial \alpha_{ij}}{\partial d_{ij}} \cdot \frac{\partial d_{ij}}{\partial \q_i}.
\label{eq_gammaij_qi}
\end{equation}

\subsubsection{C3: Collision Avoidance Constraints $\gamma_{ij}(\cdot)$}

The potential function for collision avoidance constraints is defined as
\begin{equation}
\gamma_{ij}^{*}=\left\{
\begin{aligned}
&0, &0 \le d_{ij} \le d^{\text{coll}}_{\text{min}}\\
& \frac{k_{\gamma}}{2} [1 - \cos(\frac{d_{ij} - d^{\text{coll}}_{\text{min}}}{d^{\text{coll}}_{\text{max}} - d^{\text{coll}}_{\text{min}}})\pi ], &d^{\text{coll}}_{\text{min}} < d_{ij} \le d^{\text{coll}}_{\text{max}} \\
&k_{\gamma}, &d_{ij} > d^{\text{coll}}_{\text{max}}
\end{aligned}
\right.
\label{eq_alpha_ij}
\end{equation}
Here $d^{\text{coll}}_{\text{min}}$ and $d^{\text{coll}}_{\text{max}}$ are the minimum allowed inter-robot distance and the threshold of the inter-robot distance to influence safety, respectively.

The collision avoidance constraints are different from other constraints, as the robot may fail once it collides with others.
This edge weight $\gamma_{ij}$ is defined as~\cite{robuffogiordano_PassivitybasedDecentralized_2013}:
\begin{equation}
    \gamma_{ij} = \left( \prod_{k\in \mathcal{N}_{i}} \gamma^{*}_{ik} \right) \cdot \left( \prod_{k\in \mathcal{N}_{j}/\{i\}} \gamma_{jk}^{*} \right) = \gamma_{i}\cdot \gamma_{j/i},
\label{eq_edge_weight_collision}
\end{equation}
which is the product of weights of all edges (specifically that reflect the collision avoidance constraints) connected to robot $i$ and $j$, without repetition.
With such, if robot $i$ collides with other robots, the weights of all outgoing edges will be zero, \ie, robot $i$ will be disconnected to $\G$.
Moreover, it ensures that $\gamma_{ij} = \gamma_{ji}$.

As introduced in Sec.~VII-A, we treat the closest obstacle point around a robot $i\in \mathcal{R}$ as a virtual neighbor, with a virtual index $i^{\text{obs}}$. 
The Eq.~(\ref{eq_edge_weight_collision}) is updated as:
\begin{equation}
    \gamma_{ij} = \left( \prod_{k\in \mathcal{N}_{i}\cup \{i^{\text{obs}}\}} \gamma^{*}_{ik} \right) \cdot \left( \prod_{k\in \mathcal{N}_{j} \cup  \{j^{\text{obs}}\}/\{i\}} \gamma_{jk}^{*} \right).
\label{eq_edge_weight_collision2}
\end{equation}
The derivative of $\gamma_{ij}$ w.r.t. $\q_i$ is calculated as
\begin{equation}
    \frac{\partial \gamma_{ij}}{\partial \q_{i}} = \gamma_{ij} \cdot \sum_{k\in \mathcal{N}_{i}\cup \{i^{\text{obs}}\}} \left( \frac{1}{\gamma_{ik}^{*}}\cdot \frac{\partial \alpha_{ik}^{*}}{\partial d_{ik}} \cdot \frac{\partial d_{ik}}{\partial \q_i}\right)
\label{eq_alphaij_qi}
\end{equation}

\subsection{Proof of Propositions}

\subsubsection{Proof of Proposition 1}

\begin{proof}
        To compute the derivative in Eq.~(8), each robot $i\in \mathcal{R}$ only needs local information from its one-hop neighbors $\mathcal{N}_{i}$.
        For each robot $j\in \mathcal{N}_{i}$, its pose $\langle \q_j, R_j\rangle$ and the convex hull $Conv(\mathcal{C}_{j}')$ derived from point cloud measurements need to be communicated to robot $i$. 
        Note robots do not need to share their raw point cloud, but only the flipped convex hull, which is more compact.
        The derivative of the other two weights $\alpha_{ij}$ (C1) and $\gamma_{ij}$ (C3) can also be obtained through distributed calculation, as proved in~\cite{robuffogiordano_PassivitybasedDecentralized_2013}.
        Therefore, this method can be deployed in a distributed manner. The only concern is that the graph Laplacian matrix is global information for the robot team. 
        However, it is shown in~\cite{yang_DecentralizedEstimation_2010} that both $\lambda_2$ and $v_{2i}$ can be estimated through distributed estimation. In conclusion, the connectivity force in Eq.~(8) can be calculated distributedly.
\end{proof}

\subsubsection{Proof of Proposition 2}
\begin{proof}
According to the definition of the connectivity velocity $\boldsymbol{u}^{\text{c}}_{i}$ in Eq.~(8), if $\lambda_2$ approaches $\lambda_{2}^{\text{min}}$, the term $\frac{1}{(\lambda_{2} - \lambda_{2}^{\text{min}})^{2}}$ grows unbounded.
When the navigation velocity $\boldsymbol{u}^{\text{n}}_{i}$ is bounded, the connectivity velocity $\boldsymbol{f}^{\text{c}}_i$ will dominate the movement of robot $i$, forcing robots to be connected.
Note that Prop.~2 is also verified in experiments as shown in Fig.~7.
\end{proof}

\end{document}